\documentclass[conference]{IEEEtran}
\IEEEoverridecommandlockouts

\usepackage{cite}
\usepackage{amsmath,amssymb,amsfonts}
\usepackage{algorithmic}
\usepackage{graphicx}
\usepackage{algorithm,algorithmic}
\usepackage{hyperref}
\hypersetup{hidelinks=true}
\usepackage{textcomp}
\usepackage{comment}
\usepackage{multirow}
\usepackage[dvipsnames]{xcolor}
\usepackage[table]{xcolor}
\usepackage{colortbl}
\usepackage{booktabs}
\usepackage{makecell}
\usepackage{subfig}
\usepackage{tikz}
\usepackage{pgf-pie}
\usepackage{pgfplots}
\pgfplotsset{compat=newest}
\usepgfplotslibrary{fillbetween}
\usetikzlibrary{patterns}
\usepgfplotslibrary{polar}
\usepgfplotslibrary{groupplots}
\usepackage{pgfplotstable}
\usetikzlibrary{plotmarks}
\usetikzlibrary{calc} 
\usepackage{longtable}
\usepackage{tcolorbox}
\definecolor{ash}{gray}{0.4} 

\definecolor{SciRed}{rgb}{0.84,0.37,0.00}       
\definecolor{SciBlue}{rgb}{0.00,0.45,0.70}      
\definecolor{SciGreen}{rgb}{0.00,0.62,0.45}     
\definecolor{SciOrange}{rgb}{0.90,0.60,0.00}    
\definecolor{SciPurple}{rgb}{0.80,0.47,0.65}    
\definecolor{SciGray}{rgb}{0.35,0.35,0.35}      
\definecolor{CustomPurple}{rgb}{0.4196,0.2980,0.6039}

\definecolor{llamaBlue}{HTML}{0072B2}
\definecolor{qwenOrange}{HTML}{D55E00}
\definecolor{mistralGreen}{HTML}{009E73}
\definecolor{baselineBlack}{HTML}{000000}
\definecolor{ablationPurple}{HTML}{CC79A7}

\definecolor{posStrong}{RGB}{198,239,206}
\definecolor{posMedium}{RGB}{217,234,211}
\definecolor{posLight}{RGB}{235,245,234}

\definecolor{negStrong}{RGB}{244,204,204}
\definecolor{negMedium}{RGB}{234,209,209}
\definecolor{negLight}{RGB}{245,235,235}

\definecolor{heatgood}{HTML}{B7E4C7} 
\definecolor{heatbad}{HTML}{F2B6B6}  

\newcommand{\GoodMin}{-0.206}

\newcommand{\BadLow}{-0.09}
\newcommand{\BadHigh}{0.2}

\newcommand{\MinTint}{10}

\newcommand{\hm}[2]{%
  \begingroup
  \ifdim #1pt < \BadLow pt
    \pgfmathtruncatemacro{\pct}{%
      min(100,round(\MinTint + (100-\MinTint)*((\BadLow - #1)/(\BadLow - \GoodMin))))%
    }%
    \edef\heatcolor{heatgood!\pct!white}%
  \else
    \pgfmathtruncatemacro{\pct}{%
      min(100,max(0,round(\MinTint + (100-\MinTint)*((#1 - \BadLow)/(\BadHigh - \BadLow)))))%
    }%
    \edef\heatcolor{heatbad!\pct!white}%
  \fi
  \expandafter\cellcolor\expandafter{\heatcolor}%
  \ensuremath{#1^{#2}}%
  \endgroup
}

\pgfplotsset{
  colormap={RedWhiteGreen}{
    rgb255=(178,24,43)
    rgb255=(255,255,255)
    rgb255=(0,150,0)
  }
}

\newcommand{\sigbar}[4]{%
\draw[thick]
  (axis cs:#1-0.18,#2) -- (axis cs:#1-0.18,#2+#3)
  -- (axis cs:#1+0.18,#2+#3)
  -- (axis cs:#1+0.18,#2);
\node[font=\scriptsize]
  at (axis cs:#1,#2+#3+0.035) {#4};
}

\def\BibTeX{{\rm B\kern-.05em{\sc i\kern-.025em b}\kern-.08em
    T\kern-.1667em\lower.7ex\hbox{E}\kern-.125emX}}
\begin{document}
\title{LLM4CKD: Large Language Models for Early Stage Chronic Kidney Disease Screening}


\author{\IEEEauthorblockN{Muhammad Ashad Kabir}
\IEEEauthorblockA{\textit{School of Computing, Mathematics and Engineering} \\
\textit{Charles Sturt University}\\
NSW, Australia\\
akabir@csu.edu.au}
\and
\IEEEauthorblockN{Sirajam Munira}
\IEEEauthorblockA{\textit{Department of Computer Science} \\
\textit{Rensselaer Polytechnic Institute}\\
NY, USA \\
munirs@rpi.edu}
}

\maketitle

\begingroup
\renewcommand{\thefootnote}{}
\footnotetext{\textit{\textbf{Accepted for publication in the Proceedings of the
2026 IEEE International Conference on Data Mining (ICDM).}}}
\addtocounter{footnote}{-1}
\endgroup

\begingroup
\renewcommand{\thefootnote}{}
\footnotetext{\scriptsize
© 2026 IEEE. Personal use of this material is permitted.
Permission from IEEE must be obtained for all other uses, in any
current or future media, including reprinting/republishing this material
for advertising or promotional purposes, creating new collective works,
for resale or redistribution to servers or lists, or reuse of any
copyrighted component of this work in other works.}
\addtocounter{footnote}{-1}
\endgroup

\begin{abstract}
Early screening of chronic kidney disease (CKD) is critical for timely intervention, yet most machine learning (ML) and deep learning (DL) approaches require labeled data and model training, limiting their use in real-world screening settings. This study evaluates the effectiveness of large language models (LLMs) for CKD screening under zero-shot and few-shot in-context learning settings and compares them with traditional ML and DL methods. We propose a framework that uses clinically selected tabular features and structured prompt templates to enable LLM-based inference without task-specific training. LLM performance is evaluated across multiple prompt styles, feature configurations, and data settings, and compared with standard ML, DL, and tabular foundation model (TFM) baselines, and existing CKD screening tools. The results show that LLMs can achieve competitive performance using only a small number of examples, often matching or outperforming traditional approaches in low-data settings. However, their performance remains model-dependent and less stable as input complexity increases. In contrast, ML, DL, and TFM models show more consistent improvement with larger training data. Overall, the findings highlight a trade-off between data efficiency and stability, suggesting that LLMs may serve as a flexible complementary approach for CKD screening when labeled data are limited. 
\end{abstract}

\begin{IEEEkeywords}
Chronic Kidney Disease, Large Language Model, Few-shot, Zero-shot, Machine Learning, Deep Learning, Low-resource Healthcare.
\end{IEEEkeywords}

\section{Introduction}
\label{sec:introduction}

Chronic kidney disease (CKD) is a progressive condition characterized by sustained loss of kidney function or persistent markers of kidney damage. If undetected, CKD can progress to irreversible renal decline and end-stage renal disease, requiring costly renal replacement therapy~\cite{ckdNKF}. CKD affects approximately 850 million people worldwide, nearly 10\% of the global population~\cite{francis2024chronic}. The burden is particularly severe in low- and middle-income countries (LMICs), where nearly 80\% of individuals with CKD reside and where prevalence and mortality are higher than in high-income settings~\cite{stanifer2016chronic, george2017chronic}. In South Asia, community-based studies report high CKD prevalence and low awareness, including studies from Bangladesh showing prevalence around 22\% in rural and peri-urban populations~\cite{sarker2021community}. 

Early-stage CKD is often clinically silent, yet timely detection enables interventions such as blood pressure control, diabetes management, and nephroprotective therapy that can slow disease progression~\cite{stevens2013evaluation}. Standard screening depends on laboratory measurements, including serum creatinine, estimated glomerular filtration rate (eGFR), and albumin-to-creatinine ratio (ACR), which require diagnostic infrastructure and trained personnel~\cite{stevens2024kdigo}. These requirements limit population-level screening in rural and resource-constrained settings, where laboratory tests may be unavailable, costly, or inconsistently recorded. Consequently, there is a need for scalable, data-efficient, and context-aware CKD screening methods that can operate under heterogeneous feature availability.

Conventional machine learning (ML) and deep learning (DL) models have shown promise for CKD detection, but most rely on fixed feature spaces, well-curated datasets, and stable deployment conditions~\cite{sabanayagam2025artificial}. These assumptions are often violated in real-world clinical screening: biomarkers may be missing, feature definitions may differ across sites, and population distributions may shift. Although imputation and feature-masking methods can partially address missingness, they may reduce reliability when early-stage CKD signals are subtle~\cite{little2019statistical}. 
Prior CKD prediction studies have relied on conventional supervised ML pipelines with fixed feature sets and labeled training data~\cite{KABIR2026106435}. While effective under controlled evaluation settings, such models are less adaptable when deployed across screening contexts with missing variables, heterogeneous inputs, or shifting population characteristics.
\begin{figure*}[!t]
 \centerline{\includegraphics[width=.88\linewidth]{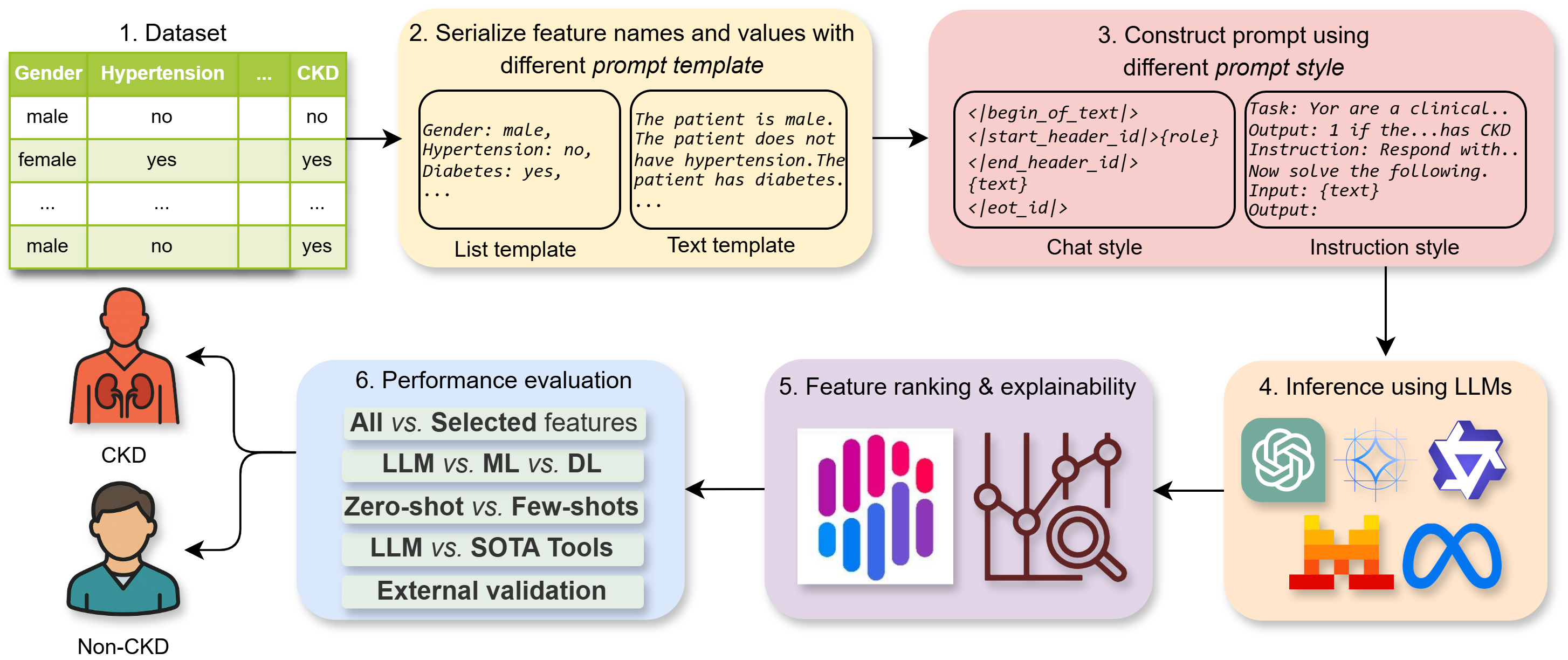}}
    \caption{Overview of the LLM4CKD evaluation framework.}
    \label{fig:methodologyLLM4CKD}
\end{figure*}

Large language models (LLMs) offer a complementary approach to CKD screening. LLMs can perform zero-shot and few-shot in-context learning~\cite{brown2020language}, reducing their reliance on large labeled datasets.
A recent study~\cite{kabir2026many} showed that feature-guided zero-shot prompting can match or improve CKD screening performance relative to full-feature inputs.
These capabilities are particularly relevant to CKD screening, where risk factors are heterogeneous and data availability varies across populations and care settings~\cite{hu2025large}. However, the extent to which LLMs can function as quantitative CKD screening models, beyond their established roles in clinical reasoning and communication, remains underexplored.

Motivated by these gaps, we propose \textit{LLM4CKD}, a framework for evaluating large language models for CKD detection in low-resource screening contexts (i.e., labeled data or diagnostic resources are limited). Our main contributions are as follows:

\begin{itemize}
    \item We present a comprehensive applied study of LLMs for low-resource early-stage CKD screening under zero-shot and few-shot in-context learning settings, and benchmark their performance against conventional ML, DL, and tabular foundation-model (TFM) baselines.

    \item We investigate the clinical plausibility of LLM decision behavior by comparing LLM-induced feature rankings with ML-based feature importance and epidemiological associations, providing insight into whether LLMs rely on clinically meaningful CKD risk factors.

     \item We compare LLMs with established CKD screening tools and conduct an independent-dataset evaluation to assess performance across datasets. 
\end{itemize}

\subsection{Related Work and Research Gaps}
\label{sec:related_work}

Existing CKD screening tools and risk scores provide useful community-level assessment mechanisms, but many were developed using cohorts from high-income countries and often emphasize later-stage disease~\cite{stolpe2022external}. This limits their applicability in LMICs, where early detection is most needed and where laboratory access, population characteristics, and healthcare pathways differ substantially. Similarly, ML- and DL-based CKD prediction studies typically assume complete or consistently measured feature sets, making them less robust to missing variables and heterogeneous inputs that are common in real-world screening environments~\cite{sabanayagam2025artificial}.

Recent work has explored LLMs across a range of nephrology and CKD-related applications, with most studies focusing on clinical support, knowledge extraction, and patient-facing communication. LLMs have been evaluated for kidney pathology support, nephrology reasoning, biopsy recommendation, and guideline-oriented decision support~\cite{syeda2024llm, toal2025large, unger2025clinical}. Other studies have applied LLMs to CKD knowledge-graph construction and feature interpretation within hybrid prediction pipelines~\cite{kumar2025multi, mishra2024kidney}, as well as patient education and other kidney-related applications~\cite{naz2024can, nurfadhilah2025retrieval}. Reviews similarly highlight their potential for summarization, decision support, education, and knowledge modeling, while noting concerns regarding reliability, consistency, and clinical validation~\cite{hu2025large, sabanayagam2025artificial}. A recent study evaluated zero-shot CKD screening from structured patient data and showed that ML-selected features generally matched or improved LLM performance relative to full-feature inputs~\cite{kabir2026many}.

Despite this progress, several gaps remain. 
First, the behavior of LLM-based CKD screening beyond zero-shot feature-guided inference, particularly under few-shot settings, remains underexplored. Second, to our knowledge, no prior study has systematically compared zero- and few-shot LLMs with conventional ML, DL, and tabular foundation-model baselines for CKD screening. Third, the alignment between LLM-derived feature importance, ML-based feature ranking, and known epidemiological risk factors has not been systematically examined. Finally, limited evidence exists on LLM-based CKD screening performance across datasets or in comparison with established CKD risk assessment tools.

\section{Methodology}
\label{sec:methodology}

Fig.~\ref{fig:methodologyLLM4CKD} provides an overview of the proposed \textit{LLM4CKD} framework for CKD screening. 
The framework comprises five key steps: (i) structured-to-text feature serialization, (ii) prompt construction, (iii) LLM-based inference, (iv) clinically grounded feature-ranking analysis, and (v) performance evaluation, including comparison with established CKD screening tools and independent-dataset evaluation to examine cross-dataset performance.
\begin{table*}[!t]
    \centering
        \caption{All features and selected features (underlined) retained after harmonization for each dataset.}
    \label{tab:datasetFeatures}
    \begin{tabular}{cp{15.5cm}}
    \hline
       Dataset  & Considered features\\
       \hline\hline
        Dataset-1~\cite{KABIR2026106435} & \underline{Age}, \underline{Gender}, \underline{Sleeping duration}, \underline{History of hypertension}, \underline{History of diabetes}, \underline{Body mass index}, \underline{Presence of red blood cells in urine}, \underline{Anemia}, \underline{Family history of hypertension}, Illiterate, Occupation, Marital status, Tobacco smoker, Smokeless tobacco, Heart disease, Stroke, Family history of diabetes, Family history of CKD, Abdominal obesity, Undernutrition, Serum albumin, Hyper-cholesterolemia, HDL cholesterol, Hypertriglyceridemia\\
        \hline
        Dataset-2~\cite{chronic_kidney_disease_336} & \underline{Age}, \underline{History of hypertension}, \underline{History of Diabetes}, \underline{Presence of red blood cells in urine}, \underline{Anemia}, Blood pressure, Specific gravity, Albumin, Sugar, Pus cell, Pus cell clumps, Bacteria, Blood glucose random, Blood urea, Sodium, Potassium, Hemoglobin, Coronary artery disease, Appetite condition, Pedal edema, White blood cell count, Red blood cell count, Packed cell volume\\\hline

        \hline
    \end{tabular}
\end{table*}
\subsection{Dataset Preparation}
\label{sec:dataset}

We evaluated \textit{LLM4CKD} using two datasets: a community-based Bangladeshi cohort for primary evaluation and an independent hospital-based Indian cohort for cross-dataset evaluation. The primary dataset, referred to as Dataset-1, was collected from the Mirzapur sub-district of Tangail, Bangladesh~\cite{sarker2021community}. It was obtained through age-stratified random sampling of adults aged 18 years and above under the Mirzapur demographic surveillance system. CKD was diagnosed using reduced kidney function, measured by estimated glomerular filtration rate (eGFR), and/or elevated urine albumin-to-creatinine ratio (uACR), according to standard clinical guidelines. To reduce label leakage, direct diagnostic markers used to define CKD status, including eGFR and uACR, were not used as model input features. After removing incomplete records, the final Dataset-1 cohort~\cite{KABIR2026106435} contained 284 participants, consisting of 112 early-stage CKD cases and 172 non-CKD individuals. The CKD cases corresponded to stages 1--3, consistent with the study objective of low-resource early-stage CKD screening. Dataset-1 contains 24 input variables (Table~\ref{tab:datasetFeatures}) covering socio-demographic characteristics, lifestyle habits, medical history, clinical examination findings, and pathology-test features.

For independent-dataset evaluation, we used Dataset-2, the UCI CKD dataset~\cite{chronic_kidney_disease_336}. Dataset-2 was collected from a hospital in India and is widely used as a benchmark for CKD detection research. It contains 400 patient records, including 250 CKD cases and 150 non-CKD controls. CKD stages are not reported in this dataset; therefore, Dataset-2 was used only for independent evaluation of cross-dataset performance, rather than early-stage-specific performance. To reduce label leakage, serum creatinine was excluded from the model inputs because it is a direct kidney-function marker strongly related to CKD diagnosis. The resulting Dataset-2 feature set contained 23 variables (Table~\ref{tab:datasetFeatures}) spanning demographic, clinical, medical-history, and non-diagnostic laboratory features.

To ensure consistency across datasets, we performed systematic feature harmonization. Variables representing the same clinical concepts were standardized by unifying feature names and value encodings into a common, semantically interpretable format suitable for LLM inference. This harmonization enabled consistent prompt construction and comparable evaluation across datasets. The complete list of harmonized features considered in this study is presented in Table~\ref{tab:datasetFeatures}. 

Feature selection was informed by prior machine learning analysis on Dataset-1~\cite{KABIR2026106435}. That analysis applied ten complementary feature selection methods, including recursive feature elimination with multiple ML estimators, to identify a compact subset of clinically meaningful and readily obtainable variables predictive of early-stage CKD. The selected features are underlined in Table~\ref{tab:datasetFeatures}. These selected Dataset-1 features were subsequently mapped to Dataset-2 by identifying variables with equivalent clinical interpretation. Because the datasets did not share identical feature spaces, only Dataset-2 variables matching the selected features from Dataset-1 were retained in the selected-feature setting.

\subsection{Feature Serialization and Prompt Style}

To enable large language models to process structured tabular data, each patient record is converted into a textual representation through feature serialization. This step maps harmonized clinical variables into text while preserving feature names, value encodings, and clinical semantics. The same serialization procedure was applied to both the all-feature and selected-feature settings to ensure comparable evaluation across prompt configurations and datasets.

Following TabLLM~\cite{hegselmann2023tabllm}, we considered two serialization styles: list-based and text-based templates. In the list-based template, each feature is represented as an explicit key--value pair, e.g., \texttt{History of hypertension: Yes; History of diabetes: No; Anemia: Yes}. This format preserves feature boundaries and provides a structured representation of each patient profile. In the text-based template, the same information is expressed as a concise clinical description, e.g., \texttt{The patient has a history of hypertension. The patient has diabetes.} This format produces a more natural narrative while retaining the underlying feature-value information.
Using both styles allows us to evaluate whether LLM-based CKD screening is sensitive to prompt serialization format.

\subsection{Prompt Construction and Templates}
\label{sec:prompt_construction}

Serialized patient records were embedded into structured prompts to evaluate LLM-based inference. We used two prompt styles: \textit{instruction-style} prompts (Table~\ref{prompt:instr-list-template}), which explicitly specify the binary CKD classification task and expected output, and \textit{chat-style} prompts (Table~\ref{prompt:chat-template}), which present the same task in a conversational format. Both styles used the same harmonized feature representations and were evaluated under all-feature and selected-feature settings.

In the zero-shot setting, prompts contained only the task instruction and the query patient record. In the few-shot setting, labeled example records were added before the query record using the same serialization format. All prompts constrained the output to one of two labels: \texttt{CKD} or \texttt{Non-CKD}.

\subsection{Large Language Models Used for Inference}
\label{sec:llm_inference_models}

We evaluated \textit{LLM4CKD} using five instruction-tuned LLMs: Gemma-2-9B~\cite{team2024gemma}, Llama-3-8B~\cite{meta2024llama3}, Qwen-3-8B~\cite{yang2025qwen3}, Mistral-7B~\cite{jiang2023mistral7b}, and GPT-4o-mini~\cite{openai2024gpt4omini}.

\captionof{table}{Instruction-style list prompt template used for inference. For the few-shot setting, labeled patient-record examples were inserted before the query patient record.}
\label{prompt:instr-list-template}
\begin{tcolorbox}
{
\textbf{Task:} You are a clinical classification model. Given a patient's structured clinical characteristics, determine whether the patient has chronic kidney disease (CKD). The target label is binary, where \texttt{1} indicates CKD and \texttt{0} indicates non-CKD.

\vspace{0.5em}
\textbf{Input:} Each patient record consists of structured clinical features, including demographic variables, medical history, laboratory measurements, and other relevant clinical attributes. 

\vspace{0.5em}
\textbf{Instruction:} Respond with exactly one token: \texttt{0} or \texttt{1}. Do not provide any explanation, justification, or additional text.

\texttt{/** Example \#1 **/}

\texttt{/** Example \#2 **/}

\texttt{/** .... **/}

\texttt{/** Example \#k **/}

\vspace{0.5em}
\textbf{Now classify the following patient.} 

History of diabetes: Yes,

History of hypertension: No,

\ldots

\textbf{Output:}
}
\end{tcolorbox}

\captionof{table}{Chat-style text prompt template for Llama.}
\label{prompt:chat-template}
\begin{tcolorbox}
{
\texttt{<|begin\_of\_text|>\textbackslash{}n}

\texttt{<|start\_header\_id|>system}

\texttt{<|end\_header\_id|>\textbackslash{}n}

You are a clinical classification model. \ldots

Each patient record consists of \ldots

Respond with exactly one token\ldots

\texttt{<|eot\_id|>\textbackslash{}n}

\texttt{<|start\_header\_id|>user}

\texttt{<|end\_header\_id|>\textbackslash{}n}

The patient is \texttt{between 40 and 49 years}. The patient is \texttt{female}. The patient is \texttt{overweight}. The patient has a family history of \texttt{hypertension}. The patient has \texttt{diabetes}. The patient has \texttt{hypertension}. The patient sleeps \texttt{less than seven hours} per day. The patient does not have \texttt{anemia}. The patient has \texttt{red blood cells in the urine}. \ldots

\texttt{<|eot\_id|>\textbackslash{}n}

\texttt{<|start\_header\_id|>assistant}

\texttt{<|end\_header\_id|>\textbackslash{}n}
}
\end{tcolorbox}

The model set includes both open-weight and proprietary LLMs, covering different model families, scales, and deployment settings. Gemma-2, Llama-3, Qwen-3, and Mistral provide compact open-weight baselines, while GPT-4o-mini provides a proprietary API-based baseline.

All models were evaluated using the same harmonized feature representations, all-feature and selected-feature settings, and zero-shot and few-shot protocols. For open-weight models, we evaluated both instruction-based and chat-style prompts. Since GPT-4o-mini is accessed through a chat-completion interface, it was evaluated only with the chat-style prompt. For probabilistic classification, token-level log-probabilities for the two constrained output labels, \texttt{0} and \texttt{1}, were converted into normalized class likelihoods, where 1 denotes CKD, and 0 denotes non-CKD.

\subsection{Feature Ranking and Explainability}
To assess interpretability and clinical plausibility, we analyze how LLMs prioritize CKD risk factors and compare their behavior with a data-driven machine learning baseline. Since LLMs do not provide explicit feature attributions for structured inputs, we use a surrogate ML model trained on the same dataset to approximate feature importance.

Feature importance is computed using SHapley Additive exPlanations (SHAP), which provides consistent, model-agnostic estimates of each feature's contribution to prediction. SHAP values are aggregated across samples to obtain global feature rankings, capturing the relative influence of clinical variables in CKD classification.
These rankings are compared with ML feature importance and known epidemiological associations. This comparison examines whether LLM-induced feature prioritization aligns with ML-based importance and clinically established CKD risk factors.

\begin{figure*}[!t]
\centering
\begin{tikzpicture}
\begin{groupplot}[
    ybar,
    enlargelimits=0.15,
    footnotesize,
    ymin=0.1, ymax=1,
    ytick={0,0.2,0.4,0.6,0.8},
    yticklabels={0,0.2,0.4,0.6,0.8},
    xtick={0,1,2,3},
    xticklabel style={rotate=45, anchor=east, align=center},
    legend columns=-1,
    nodes near coords ,
    every node near coord/.append style={rotate=90,anchor=west, /pgf/number format/precision=2,  font=\scriptsize}, 
    name=mygroup,
    group style={
        group size=5 by 1,
        xlabels at=edge bottom,
        horizontal sep=.7cm,   
        ylabels at=edge left
    }
]

\nextgroupplot[
    width=0.26\textwidth,
    height=5cm,
    bar width=6pt,
    ylabel={Balanced Accuracy},
    xlabel={(a) Gemma-2},
        xticklabels={
        Inst+List,
        \textbf{Inst+Text},
        Chat+List,
        Chat+Text
    },
    legend style = { text=black, column sep = 1pt, legend columns = -1, legend to name = grouplegend}
]

\addplot[pattern=dots, pattern color=SciRed, draw=SciRed, error bars/.cd, y dir=both, y explicit] 
    coordinates { (0, 0.7576)
    (1, 0.7659) (2, 0.7513) (3, 0.711)};
\addlegendentry{All features}
\addplot[pattern=north east lines, pattern color=SciBlue, draw=SciBlue,  error bars/.cd, 
          y dir=both, y explicit] 
    coordinates {      
        (0, 0.6536) 
        (1, 0.7299) 
        (2, 0.6033) 
        (3, 0.6404)
        };
\addlegendentry{Selected features}
\sigbar{0}{1}{0.04}{$***$} 
\sigbar{1}{1}{0.04}{$n.s.$} 
\sigbar{2}{1}{0.04}{$**$} 
\sigbar{3}{1}{0.04}{$*$} 
\nextgroupplot[
     width=0.26\textwidth,
    height=5cm,
    bar width=6pt,
    xlabel={(b)} Llama-3,
    xticklabels={
        Inst+List,
        Inst+Text,
        Chat+List,
        \textbf{Chat+Text}
    },
    ]
\addplot[pattern=dots, pattern color=SciRed, draw=SciRed,  error bars/.cd,
          y dir=both, y explicit] coordinates {(0,0.5) (1,0.4987) (2,0.5029) (3,0.5130)};
\addplot[pattern=north east lines, pattern color=SciBlue, draw=SciBlue,  error bars/.cd, 
          y dir=both, y explicit] coordinates {(0,0.5419) (1,0.6798) (2,0.645) (3,0.7422)};

\sigbar{0}{1}{0.04}{$***$} 
\sigbar{1}{1}{0.04}{$***$}
\sigbar{2}{1}{0.04}{$***$}
\sigbar{3}{1}{0.04}{$***$}

\nextgroupplot[
    width=0.26\textwidth,
    height=5cm,
    bar width=6pt,
    xlabel={(c)} Qwen-3,
    xticklabels={
    \textbf{Inst+List},
    Inst+Text,
    Chat+List,
    Chat+Text
    },
    ]
\addplot[pattern=dots, pattern color=SciRed, draw=SciRed,  error bars/.cd,
          y dir=both, y explicit] coordinates {(0,0.6206) (1,0.5181) (2,0.5) (3,0.63325)};
\addplot[pattern=north east lines, pattern color=SciBlue, draw=SciBlue,  error bars/.cd, 
          y dir=both, y explicit] coordinates {(0,0.7829) (1,0.6239) (2,0.50) (3,0.5346)};

\sigbar{0}{1}{0.04}{$***$}
\sigbar{1}{1}{0.04}{$n.s.$}
\sigbar{2}{1}{0.04}{$***$}
\sigbar{3}{1}{0.04}{$***$}

\nextgroupplot[
    width=0.26\textwidth,
    height=5cm,
    bar width=6pt,
    xticklabels={
    Inst+List,
    \textbf{Inst+Text},
    Chat+List,
    Chat+Text
    },
    xlabel={(d)} Mistral]
\addplot[pattern=dots, pattern color=SciRed, draw=SciRed,  error bars/.cd,
          y dir=both, y explicit] coordinates {(0,0.5552) (1,0.7231) (2,0.67198) (3,0.61315)};
\addplot[pattern=north east lines, pattern color=SciBlue, draw=SciBlue,  error bars/.cd, 
          y dir=both, y explicit] coordinates {(0,0.7721) (1,0.7925) (2,0.52711) (3,0.79346)};
\sigbar{0}{1}{0.04}{$***$}
\sigbar{1}{1}{0.04}{$***$}
\sigbar{2}{1}{0.04}{$***$}
\sigbar{3}{1}{0.04}{$***$}

\nextgroupplot[
    width=3cm,
    height=5cm,
    bar width=6pt,
    enlarge x limits=0.4,
    xlabel={(e)} GPT-4o-mini,
    xticklabels={
    Inst+List,
    Inst+Text,
    \textbf{Chat+List},
    Chat+Text
    },
    ]
\addplot[pattern=dots, pattern color=SciRed, draw=SciRed,  error bars/.cd,
          y dir=both, y explicit] coordinates { (2,0.6972) (3,0.7542)};
\addplot[pattern=north east lines, pattern color=SciBlue, draw=SciBlue,  error bars/.cd, 
          y dir=both, y explicit] coordinates {(2,0.7708) (3,0.7021)};
\sigbar{2}{1}{0.04}{$***$}
\sigbar{3}{1}{0.04}{$n.s.$} 
\end{groupplot}
\node at ($(current bounding box.north) + (0cm, 0.6cm)$) {\ref{grouplegend}};

\end{tikzpicture}
\caption{Zero-shot comparison of all versus selected features and the performance of different prompt styles (instruction vs. chat) and templates (list vs. text) for LLMs. Bars report balanced accuracy, while statistical significance between all- and selected-feature settings is assessed using a paired permutation test on per-sample Brier loss. Significance levels are indicated by asterisks: *$p<0.05$, **$p<0.01$ and ***$p<0.001$. Bold denotes the best-performing prompt style and template.} 
\label{fig:allvsS1}
\end{figure*}
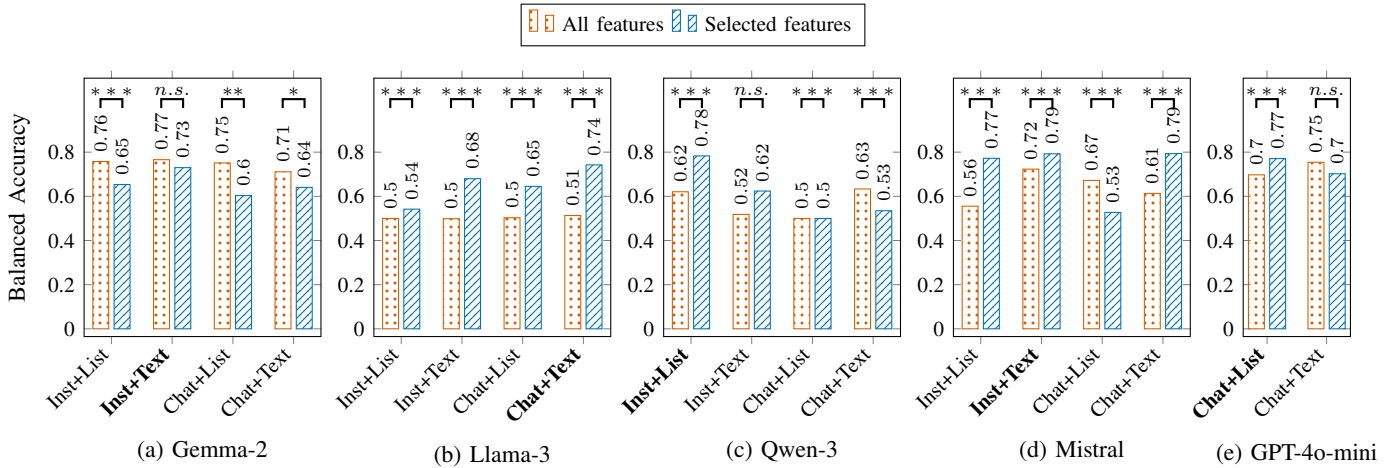

\subsection{Model Evaluation}
\label{sec:model_evaluation}

For zero-shot LLM evaluation, models were evaluated without labeled examples, training, or parameter updates. Since zero-shot inference does not require a training split, each model was evaluated directly on the full dataset.

For few-shot LLM evaluation and supervised baselines, each dataset was partitioned into 80\% training and 20\% testing using stratified sampling. The 20\% held-out test partition was consistently used for final evaluation across all models. Few-shot in-context examples for LLMs and low-data training subsets of size 4, 8, 16, and 32 for supervised ML, DL, and tabular foundation-model (TFM) baselines were sampled only from the 80\% training partition.

To account for sampling variability, each configuration was repeated across five random seeds: \{0, 1, 32, 42, 1024\}. For each seed, the same train--test partitioning and sampled training examples were used across all models to ensure paired comparison. We report mean performance across runs using balanced accuracy and probabilistic metrics, including Brier loss. Statistical significance between model pairs was assessed using paired permutation tests and mixed-effects analysis on per-sample Brier loss across repeated runs.

For LLM inference, open-weight models were executed locally with temperature set to 0 to ensure deterministic decoding. The pretrained weights of Gemma-2-9B\footnote{\url{https://huggingface.co/google/gemma-2-9b-it}}, Llama-3-8B\footnote{\url{https://huggingface.co/meta-llama/Meta-Llama-3-8B-Instruct}}, Qwen-3-8B\footnote{\url{https://huggingface.co/Qwen/Qwen3-8B}}, and Mistral-7B\footnote{\url{https://huggingface.co/mistralai/Mistral-7B-Instruct-v0.3}} were obtained from Hugging Face. GPT-4o-mini was accessed through the OpenAI API. For probabilistic classification, we computed normalized class probabilities from output token log-probabilities over the predefined label tokens corresponding to CKD and non-CKD. These normalized probabilities were used to derive predicted labels, Brier loss, and confidence scores.

For conventional ML baselines, models were implemented using scikit-learn. Unless otherwise specified, default hyperparameters were used, with \texttt{class\_weight='balanced'} applied where supported to account for class imbalance. DL and TFM baselines were implemented using their standard public implementations: TabNet was implemented in PyTorch, NODE was implemented using the official repository,\footnote{\url{https://github.com/Qwicen/node}} TabPFN was installed using the \texttt{tabpfn} Python package, and SAINT was implemented using the official repository.\footnote{\url{https://github.com/somepago/saint}} Model-internal random states were fixed at 42 where applicable, while data partitioning and subset sampling followed the five evaluation seeds above.

The source code for our experimental pipeline is publicly available.\footnote{\url{https://github.com/akabircs/LLM4CKD}} Experiments were conducted on a workstation equipped with an NVIDIA GeForce RTX 3080 16GB GPU.

\begin{figure*}
    \centering
\begin{tikzpicture}
\begin{groupplot}[
    group style={
        group size= 3 by 1,
        horizontal sep=1.2cm,
        vertical sep=2.4cm,
    },
]

\nextgroupplot[xlabel={\shortstack{Shots\\(a) LLM}},
    width=0.32\textwidth,
    height=5cm,
    ylabel={Balanced accuracy},
    ymin=0.6, ymax=0.85,
    symbolic x coords={0,4,8,16,32},
    xtick=data,
    legend style={at={(0.5,1.45)},font = \footnotesize, anchor=north,legend columns=2}
]
\addplot[color=SciRed, mark=*,thick]
coordinates {(0,0.7101) (4,0.8029) (8,0.7614) (16,0.7562) (32,0.7873)};
\addlegendentry{Gemma-2}

\addplot[color=SciOrange,mark=+,thick]
coordinates {(0,0.7857) (4,0.8044) (8,0.8138) (16,0.8042) (32,0.8070)};
\addlegendentry{GPT-4o-mini}

\addplot[color=cyan,mark=x, thick]
coordinates {(0,0.7516) (4,0.6827) (8,0.6745) (16,0.6249) (32,0.7392)};
\addlegendentry{Llama-3}

\addplot[color=SciGreen, mark=diamond*, thick]
coordinates {(0,0.7781) (4,0.8135) (8,0.7117) (16,0.7825) (32,0.7639)};
\addlegendentry{Qwen-3}

\addplot[ color=gray, mark=triangle*,thick]
coordinates {(0,0.7591) (4,0.7494) (8,0.7336) (16,0.7169) (32,0.6773)};
\addlegendentry{Mistral}

\nextgroupplot[xlabel={\shortstack{Training samples\\(b) Machine learning}},
    width=0.32\textwidth,
    height=5cm,
    symbolic x coords={4,8,16,32},
    xtick=data,
    legend style={at={(0.5,1.45)},font = \footnotesize, anchor=north,legend columns=3}
]
\addplot[color=SciRed,mark=*,thick]
coordinates {(4,0.7203) (8,0.7270) (16,0.7678) (32,0.7648)};
\addlegendentry{RF}

\addplot[color=SciOrange, mark=+,thick]
coordinates {(4,0.6577) (8,0.6619) (16,0.7222) (32,0.7244)};
\addlegendentry{GB}

\addplot[color=cyan,mark=x,thick]
coordinates {(4,0.7201) (8,0.7377) (16,0.7478) (32,0.7636)};
\addlegendentry{ET}

\addplot[color=SciGreen,mark=diamond*,thick]
coordinates {(4,0.7049) (8,0.7281) (16,0.7592) (32,0.7747)};
\addlegendentry{LR}

\addplot[color=gray, mark=triangle*, thick]
coordinates {(4,0.6844) (8,0.6886) (16,0.6965) (32,0.7751)};
\addlegendentry{AB}

\addplot[color=SciPurple, mark=star, thick]
coordinates {(4,0.6735) (8,0.7206) (16,0.7114) (32,0.7375)};
\addlegendentry{DT}

\addplot[color=Tan, mark=square*,thick]
coordinates {(4,0.6577) (8,0.6619) (16,0.7222) (32,0.7244)};
\addlegendentry{XGB}

\addplot[color=SciBlue, mark=pentagon*,thick]
coordinates {(4,0.7542) (8,0.7523) (16,0.7551) (32,0.771) };
\addlegendentry{MLP}

\addplot[color=CustomPurple,mark=oplus,thick]
coordinates {(4,0.651177) (8,0.6619) (16,0.7222) (32,0.7244)};
\addlegendentry{LGB}

\nextgroupplot[xlabel={\shortstack{Training/context samples\\(c) Deep learning/TFM}},
    width=0.32\textwidth,
    height=5cm,
    symbolic x coords={4,8,16,32},
    xtick=data,
    legend style={at={(0.5,1.45)},font = \footnotesize, anchor=north,legend columns=2}
]
\addplot[color=SciBlue, mark=pentagon*, thick]
coordinates {(4,0.6582) (8,0.7508) (16,0.7616) (32,0.8196) };
\addlegendentry{TabPFN}

\addplot[color=CustomPurple,mark=oplus,thick]
coordinates {(4,0.7677) (8,0.7530) (16,0.7836) (32,0.7974) };
\addlegendentry{NODE}

\addplot[color=Maroon, mark=+, thick]
coordinates {(4,0.7304) (8,0.7495)  (16,0.7591)  (32,0.7329)};
\addlegendentry{SAINT}

\addplot[color=Tan,mark=square*, thick]
coordinates {(4,0.5495) (8,0.5997) (16,0.7160) (32,0.7136) };
\addlegendentry{TabNet}

\end{groupplot}
\end{tikzpicture}
    \caption{Balanced accuracy comparison using selected features across (a) LLMs, (b) ML models, and (c) DL/TFM models. LLM performance is evaluated using varying numbers of in-context examples (shots), while supervised ML and DL baselines are trained on corresponding labeled subsets and TabPFN uses them as labeled context samples. Results are averaged over five runs with different random seeds; within each run, identical training/in-context and test samples are used across all models.}
    \label{fig:llmvsmlvsdl}
\end{figure*}
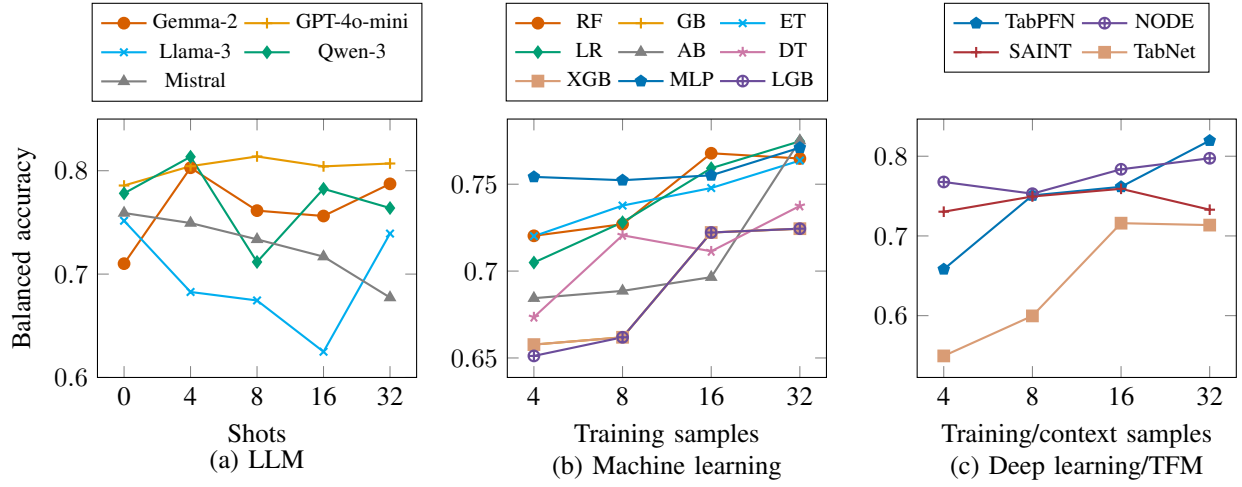

\section{Results and Discussion}
\subsection{Zero-Shot Analysis of Feature Selection and Prompt Design}
\label{sec:all_selected_features}

Fig.~\ref{fig:allvsS1} compares all-feature and selected-feature inputs under the zero-shot setting across prompt styles and serialization templates for each LLM. No labeled examples, training data, or parameter updates were used in this experiment. Performance is reported using balanced accuracy, while statistical significance between paired all-feature and selected-feature predictions is assessed using a paired permutation test on per-sample Brier loss over the same samples. This comparison evaluates whether reducing input complexity improves zero-shot LLM-based CKD screening.

Overall, selected features improve or maintain performance for most LLMs, but the effect remains model- and prompt-dependent. Llama-3 benefits most consistently, with selected-feature inputs improving balanced accuracy across all prompt-template combinations. The largest gain occurs in the Chat+Text setting, where balanced accuracy increases from 0.51 to 0.74. Mistral also benefits in most settings, especially under instruction-style prompts, increasing from 0.56 to 0.77 with Instruction+List and from 0.72 to 0.79 with Instruction+Text. However, its Chat+List performance drops from 0.67 to 0.50, indicating that feature selection alone is insufficient without an effective prompt format.

Qwen-3 shows mixed behavior. Selected features improve Instruction+List from 0.62 to 0.78 and Instruction+Text from 0.52 to 0.62, while Chat+List remains at 0.50 and Chat+Text decreases from 0.63 to 0.53. In contrast, the Instruction+Text setting changes only slightly, and the Chat+List setting remains near chance-level balanced accuracy. GPT-4o-mini, evaluated only with chat-style prompts, shows stable behavior: selected features improve Chat+List from 0.70 to 0.77 but slightly reduce Chat+Text from 0.75 to 0.70.

Gemma-2 differs from the other models. Selected features reduce balanced accuracy in most prompt-template settings, including Instruction+List, Chat+List, and Chat+Text, while Instruction+Text remains relatively stable. This suggests that Gemma-2 may rely more on broader contextual information from the full feature set, whereas other models often benefit from compact, clinically relevant inputs.

These findings show that selected-feature prompting reduces input complexity and improves zero-shot LLM-based CKD screening, particularly for Llama-3, Mistral, Qwen-3, and GPT-4o-mini. However, the gains are not universal: performance depends jointly on the LLM, prompt style, and serialization template. Instruction-style prompts are generally effective for open-weight models, while chat-style prompting remains necessary for GPT-4o-mini. These results highlight the importance of evaluating feature selection and prompt design together rather than treating them as independent choices.

\subsection{Low-Data Comparison with ML, DL, and Tabular Foundation Models}
\label{sec:low_data_comparison}
Fig.~\ref{fig:llmvsmlvsdl} compares LLMs with conventional ML and tabular DL and foundation-model baselines under limited labeled-data settings using the selected-feature set. LLMs are evaluated using zero-shot and few-shot in-context learning, whereas ML and tabular baselines are trained with progressively larger labeled subsets. For each seed, the same selected features, training, and in-context examples, and held-out test samples are used across models. 

LLMs achieve competitive balanced accuracy with very few examples, but their performance does not consistently improve as the number of shots increases. Gemma-2, Qwen-3, and GPT-4o-mini improve from zero-shot to four-shot inference, reaching approximately 0.80--0.81 balanced accuracy. In contrast, Llama-3 and Mistral show weaker or declining performance as more examples are added, indicating sensitivity to in-context example composition. GPT-4o-mini is the most stable LLM, maintaining balanced accuracy around 0.80 across shots. 

Supervised ML models show a more predictable trend: performance generally improves as the number of labeled training samples increases. Random Forest, Logistic Regression, Extra Trees, XGBoost, and MLP improve or remain competitive as the training size grows from 4 to 32 samples. Tabular DL and foundation models show a similar scaling pattern, with TabPFN and NODE achieving the strongest performance at larger training sizes. 

Overall, the results show a trade-off between data efficiency and scalability under the selected-feature setting. LLMs can provide strong performance with only a few in-context examples, making them attractive for low-resource CKD screening when labeled data are scarce. However, additional shots do not guarantee improvement. In contrast, supervised ML and tabular models require labeled training data but tend to benefit more consistently as data availability increases.

\subsection{Pairwise Brier Loss Comparison}
\label{sec:pairwise_loss}

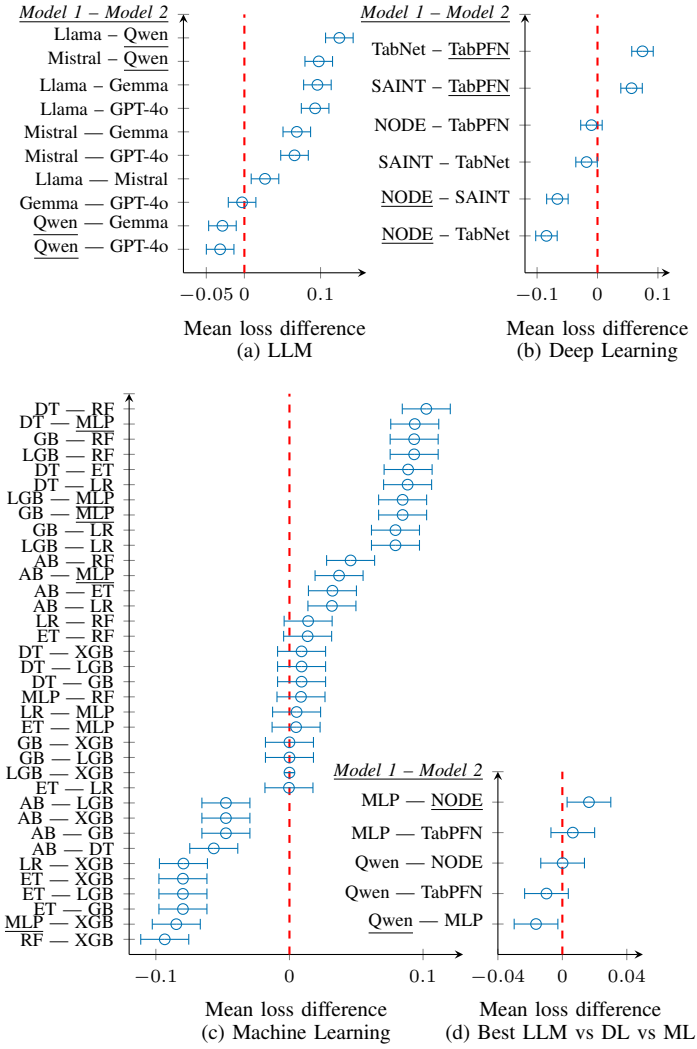
\begin{figure}[!t]
    \centering
\begin{tikzpicture}
\begin{groupplot}[
    name=topgroup,
    group style={
        group size= 2 by 1,
        horizontal sep=2.1cm,
    },
]

\nextgroupplot[
    xmin=-0.08, xmax=0.16,
    xshift=-9cm,
    xlabel style={font=\footnotesize},
    xlabel={\shortstack{Mean loss difference\\(a) LLM}},
    height=5cm,
    width=4cm,
    axis y line=left,
    xticklabel style={font=\scriptsize},
    yticklabel style={font=\scriptsize},
    ytick={10,9,8,7,6,5,4,3,2,1,11},
    yticklabels={
        Llama -- \underline{Qwen},
        Mistral -- \underline{Qwen},
        Llama -- Gemma,
        Llama -- GPT-4o,
        Mistral –- Gemma,
        Mistral –- GPT-4o,
        Llama –- Mistral,
        Gemma –- GPT-4o,
        \underline{Qwen} –- Gemma,
        \underline{Qwen} –- GPT-4o,
        \underline{\textit{Model 1 -- Model 2}},
    },
    xtick={-0.05,0,0.1},
    scaled x ticks=false,
    x tick label style={
        /pgf/number format/fixed,
        /pgf/number format/precision=2
    },
    axis x line=bottom,
]

\addplot[red, dashed, thick] coordinates {(0,0) (0,11)};

\addplot+[
    only marks,
    mark=o,
    SciBlue,
    error bars/.cd,
        x dir=both,
        x explicit
] coordinates {
(0.1246918144041334,10) +- (0.018104667903794735,0)
(0.09762796278416112,9) +- (0.018104667903794704,0)
(0.09587910447107013,8) +- (0.01810466790379474,0)
(0.09287619947517758,7) +- (0.018104667903794728,0)
(0.06881525285109782,6) +- (0.018104667903794704,0)
(0.06581234785520527,5) +- (0.018104667903794704,0)
(0.02706385161997231,4) +- (0.01810466790379474,0)
(-0.00300290499589255,3) +- (0.01810466790379471,0)
(-0.0288127099330633,2) +- (0.018104667903794707,0)
(-0.03181561492895585,1) +- (0.0181046679037947,0)
};

\nextgroupplot[
    xmin=-0.12, xmax=0.12,
    xlabel style={font=\footnotesize},
    xlabel={\shortstack{Mean loss difference\\(b) Deep Learning}},
    height=5cm,
    width=3.5cm,
    axis y line=left,
    xticklabel style={font=\scriptsize},
    yticklabel style={font=\scriptsize},
    ytick={1,2,3,4,5,6,7},
    yticklabels={
        \underline{NODE} -- TabNet,
        \underline{NODE} -- SAINT,
        SAINT -- TabNet,
        NODE -- TabPFN,
        SAINT -- \underline{TabPFN},
        TabNet -- \underline{TabPFN},
           \underline{\textit{Model 1 -- Model 2}},
    },
    xtick={-0.1,0,0.1},
    scaled x ticks=false,
    x tick label style={
        /pgf/number format/fixed,
        /pgf/number format/precision=2
    },
    axis x line=bottom,
]

\addplot[red, dashed, thick] coordinates {(0,0) (0,7)};

\addplot+[
    only marks,
    mark=o,
    SciBlue,
    error bars/.cd,
        x dir=both,
        x explicit
] coordinates {
    (-0.08448698692156771,1) +- (0.017779345224478563,0)
    (-0.06637828835567144,2) +- (0.017779345224478563,0)
    (-0.01810869856589627,3) +- (0.017779345224478577,0)
    (-0.01002451593664392,4) +- (0.017779345224478573,0)
    (0.05635377241902752,5) +- (0.01777934522447857,0)
    (0.07446247098492378,6) +- (0.017779345224478577,0)
};
\end{groupplot}
\hspace{-3cm}
\coordinate (rowoneSW) at (current bounding box.north west);

\begin{groupplot}[
at={(rowoneSW)},
  anchor=north east,
    group style={
        group size= 2 by 1,
    },
]
\nextgroupplot[
    xmin=-0.12, xmax=0.13,
    xlabel style={font=\footnotesize},
    xlabel={\shortstack{Mean loss difference\\(c) Machine Learning}},
    height=9cm,
    yshift=-1.6cm,
    xshift = -2.3cm,
    width=6cm,
    axis y line=left,
    xticklabel style={font=\scriptsize},
    yticklabel style={font=\scriptsize},
    ytick={1,...,37},
    yticklabels={
RF –- XGB,
\underline{MLP} –- XGB,
ET –- GB,
ET –- LGB,
ET –- XGB,
LR –- XGB,
AB –- DT,
AB –- GB,
AB –- XGB,
AB –- LGB,
ET –- LR,
LGB –- XGB,
GB –- LGB,
GB –- XGB,
ET –- MLP,
LR –- MLP,
MLP –- RF,
DT –- GB,
DT –- LGB,
DT –- XGB,
ET –- RF,
LR –- RF,
AB –- LR,
AB –- ET,
AB –- \underline{MLP},
AB –- RF,
LGB –- LR,
GB –- LR,
GB –- \underline{MLP},
LGB –- \underline{MLP},
DT –- LR,
DT –- ET,
LGB –- RF,
GB –- RF,
DT –- \underline{MLP},
DT –- RF,
    },
    xtick={-0.1,0,0.1},
    scaled x ticks=false,
    x tick label style={
        /pgf/number format/fixed,
        /pgf/number format/precision=2
    },
    axis x line=bottom,
]

\addplot[red, dashed, thick] coordinates {(0,0) (0,37)};

\addplot+[
    only marks,
    mark=o,
    SciBlue,
    error bars/.cd,
        x dir=both,
        x explicit
] coordinates {
(-0.09334678053402812,1) +- (0.017966696006685517,0)
(-0.08470933134837164,2) +- (0.017966696006685524,0)
(-0.07973667539955398,3) +- (0.017966696006685528,0)
(-0.07973667539955398,4) +- (0.017966696006685528,0)
(-0.07973667539955398,5) +- (0.017966696006685528,0)
(-0.07936372269805758,6) +- (0.017966696006685528,0)
(-0.05664946165432394,7) +- (0.017966696006685455,0)
(-0.04756987730728964,8) +- (0.017966696006685455,0)
(-0.04756987730728963,9) +- (0.017966696006685448,0)
(-0.04756987730728963,10) +- (0.017966696006685448,0)
(-0.0003729527014964054,11) +- (0.01796669600668553,0)
(0.0,12) +- (0.0,0)
(6.938893903907228e-18,13) +- (0.017966696006685524,0)
(6.938893903907228e-18,14) +- (0.017966696006685524,0)
(0.004972655948817649,15) +- (0.01796669600668553,0)
(0.005345608650314054,16) +- (0.01796669600668553,0)
(0.008637449185656489,17) +- (0.017966696006685524,0)
(0.009079584347034299,18) +- (0.017966696006685528,0)
(0.009079584347034306,19) +- (0.017966696006685528,0)
(0.009079584347034306,20) +- (0.017966696006685528,0)
(0.01361010513447414,21) +- (0.017966696006685535,0)
(0.01398305783597054,22) +- (0.01796669600668552,0)
(0.03179384539076794,23) +- (0.017966696006685455,0)
(0.03216679809226435,24) +- (0.01796669600668546,0)
(0.03713945404108199,25) +- (0.017966696006685448,0)
(0.04577690322673848,26) +- (0.01796669600668545,0)
(0.07936372269805758,27) +- (0.017966696006685528,0)
(0.07936372269805758,28) +- (0.017966696006685524,0)
(0.08470933134837164,29) +- (0.017966696006685524,0)
(0.08470933134837164,30) +- (0.017966696006685524,0)
(0.08844330704509187,31) +- (0.01796669600668553,0)
(0.08881625974658829,32) +- (0.017966696006685535,0)
(0.09334678053402812,33) +- (0.017966696006685517,0)
(0.09334678053402812,34) +- (0.017966696006685517,0)
(0.09378891569540593,35) +- (0.017966696006685528,0)
(0.1024263648810624,36) +- (0.01796669600668553,0)
};
\nextgroupplot[
        xmin=-0.04, xmax=0.05,
    xlabel style={font=\footnotesize},
    xlabel={\shortstack{Mean loss difference\\(d) Best LLM vs DL vs ML}},
    height=4cm,
    width=3.5cm,
    axis y line=left,
    yshift = -2.5cm,
    xshift = -.54cm,
    xticklabel style={font=\scriptsize},
    yticklabel style={font=\scriptsize},
    ytick={1,2,3,4,5,6},
    yticklabels={
        \underline{Qwen} –- MLP,
        Qwen –- TabPFN,
        Qwen –- NODE,
        MLP –- TabPFN,
        MLP –- \underline{NODE},
           \underline{\textit{Model 1 -- Model 2}},
    },
    xtick={-0.04,0,0.04},
    scaled x ticks=false,
    x tick label style={
        /pgf/number format/fixed,
        /pgf/number format/precision=2
    },
    axis x line=bottom,
]
\addplot[red, dashed, thick] coordinates {(0,0) (0,6)};

\addplot+[
    only marks,
    mark=o,
    SciBlue,
    error bars/.cd,
        x dir=both,
        x explicit
] coordinates {
    (-0.016342664455136987,1) +- (0.01355782304522301,0)
    (-0.009859019270995195,2) +- (0.01355782304522301,0)
    (0.0001654966656488808,3) +- (0.013557823045223008,0)
    (0.006483645184141792,4) +- (0.013557823045223051,0)
    (0.016508161120785868,5) +- (0.013557823045223051,0)
};
\end{groupplot}
\end{tikzpicture}
    \caption{Pairwise comparison of mean Brier loss differences across model families: (a) LLMs, (b) DL models, (c) ML models, and (d) best-performing LLMs versus DL versus ML models. Each point denotes the mean loss difference between two models (Model 1 $-$ Model 2), with horizontal error bars showing 95\% confidence intervals estimated from a mixed-effects model accounting for repeated stochastic evaluations. The vertical red dashed line indicates zero difference. Negative values indicate that Model 1 outperforms Model 2 (lower mean loss), while positive values indicate the opposite. Differences are considered statistically significant when the confidence interval does not intersect the zero line. Underlined model names denote the best-performing model(s) within their group.}

    \label{fig:meanloss}
\end{figure}

Fig.~\ref{fig:meanloss} reports pairwise mean Brier loss differences across LLM, ML, and tabular DL and foundation-model groups. For each comparison, the plotted value is $\Delta = \text{Loss}(\text{Model 1})-\text{Loss}(\text{Model 2})$; therefore, negative values favor Model 1 and positive values favor Model 2. Confidence intervals that do not cross zero indicate statistically significant differences under the mixed-effects model.

Within the LLM group, Qwen-3 shows the strongest overall probabilistic performance, achieving lower Brier loss than Llama-3 and Mistral while remaining competitive with Gemma-2 and GPT-4o-mini. This finding is consistent with Table~\ref{tab:top4_metrics}, where Qwen-3 obtains strong Brier loss across shot settings, including 0.154 at four shots, 0.141 at 16 shots, and 0.146 at 32 shots. Qwen-3 also achieves the highest four-shot balanced accuracy among the top models (0.814) and high sensitivity (0.873), indicating strong early-stage CKD case detection in the extreme low-data regime. GPT-4o-mini and Gemma-2 form a competitive middle tier, while Llama-3 and Mistral generally show higher loss.

Among tabular DL and foundation models, NODE and TabPFN provide the strongest performance, with lower loss than SAINT and TabNet in most pairwise comparisons. Table~\ref{tab:top4_metrics} further shows that their advantage becomes clearer as the number of labeled training samples increases. At 32 context samples, TabPFN achieves the best overall balanced accuracy among the top models (0.820) and the lowest Brier loss (0.122), while NODE also remains strong with balanced accuracy of 0.797 and Brier loss of 0.142. These results suggest that tabular foundation models benefit more consistently from additional labeled data than LLMs, particularly in probabilistic calibration.

\begin{table}[!t]
    \centering
\caption{Detailed performance comparison of the top representative LLM, ML, and tabular DL and foundation-model baselines using selected features. Metrics include balanced accuracy, AUROC, macro-F1, sensitivity, Brier loss, and expected calibration error (ECE). Values are reported as means across five random-seed iterations, with standard deviations as subscripts and bias-corrected and accelerated (BCa) 95\% confidence intervals in brackets.}
\label{tab:top4_metrics}

 \setlength{\tabcolsep}{1pt} 
    \resizebox{\linewidth}{!}{
    \begin{tabular}{c l cccccc}
    \hline
     Shot & Model & B. accuracy ($\uparrow$) & AUROC ($\uparrow$) & F1 ($\uparrow$) & Sensitivity ($\uparrow$) & Brier ($\downarrow$) & ECE ($\downarrow$) \\
     \hline\hline
     4 & Qwen & \makecell[t]{$0.814_{\textcolor{ash}{\pm 0.025}}$\\{\scriptsize \textcolor{ash}{[0.763, 0.854]}}} &\makecell[t]{$0.867_{\textcolor{ash}{\pm 0.032}}$\\{\scriptsize \textcolor{ash}{[0.813, 0.907]}}} &\makecell[t]{$0.796_{\textcolor{ash}{\pm 0.024}}$\\{\scriptsize \textcolor{ash}{[0.747, 0.842]}}} &\makecell[t]{$0.873_{\textcolor{ash}{\pm 0.067}}$\\{\scriptsize \textcolor{ash}{[0.799, 0.925]}}} &\makecell[t]{$0.154_{\textcolor{ash}{\pm 0.022}}$\\{\scriptsize \textcolor{ash}{[0.128, 0.185]}}} &\makecell[t]{$0.095_{\textcolor{ash}{\pm 0.054}}$\\{\scriptsize \textcolor{ash}{[0.071, 0.097]}}} \\
 & TabPFN & \makecell[t]{$0.658_{\textcolor{ash}{\pm 0.105}}$\\{\scriptsize \textcolor{ash}{[0.609, 0.701]}}} &\makecell[t]{$0.813_{\textcolor{ash}{\pm 0.051}}$\\{\scriptsize \textcolor{ash}{[0.748, 0.865]}}} &\makecell[t]{$0.656_{\textcolor{ash}{\pm 0.116}}$\\{\scriptsize \textcolor{ash}{[0.604, 0.710]}}} &\makecell[t]{$0.436_{\textcolor{ash}{\pm 0.186}}$\\{\scriptsize \textcolor{ash}{[0.357, 0.513]}}} &\makecell[t]{$0.176_{\textcolor{ash}{\pm 0.032}}$\\{\scriptsize \textcolor{ash}{[0.155, 0.200]}}} &\makecell[t]{$0.096_{\textcolor{ash}{\pm 0.037}}$\\{\scriptsize \textcolor{ash}{[0.064, 0.106]}}} \\
 & NODE & \makecell[t]{$0.768_{\textcolor{ash}{\pm 0.053}}$\\{\scriptsize \textcolor{ash}{[0.711, 0.811]}}} &\makecell[t]{$0.863_{\textcolor{ash}{\pm 0.036}}$\\{\scriptsize \textcolor{ash}{[0.810, 0.903]}}} &\makecell[t]{$0.764_{\textcolor{ash}{\pm 0.043}}$\\{\scriptsize \textcolor{ash}{[0.708, 0.813]}}} &\makecell[t]{$0.718_{\textcolor{ash}{\pm 0.152}}$\\{\scriptsize \textcolor{ash}{[0.635, 0.790]}}} &\makecell[t]{$0.160_{\textcolor{ash}{\pm 0.017}}$\\{\scriptsize \textcolor{ash}{[0.132, 0.194]}}} &\makecell[t]{$0.118_{\textcolor{ash}{\pm 0.026}}$\\{\scriptsize \textcolor{ash}{[0.068, 0.132]}}} \\
 & MLP & \makecell[t]{$0.754_{\textcolor{ash}{\pm 0.068}}$\\{\scriptsize \textcolor{ash}{[0.712, 0.798]}}} &\makecell[t]{$0.827_{\textcolor{ash}{\pm 0.061}}$\\{\scriptsize \textcolor{ash}{[0.766, 0.877]}}} &\makecell[t]{$0.757_{\textcolor{ash}{\pm 0.069}}$\\{\scriptsize \textcolor{ash}{[0.714, 0.806]}}} &\makecell[t]{$0.645_{\textcolor{ash}{\pm 0.142}}$\\{\scriptsize \textcolor{ash}{[0.574, 0.717]}}} &\makecell[t]{$0.161_{\textcolor{ash}{\pm 0.030}}$\\{\scriptsize \textcolor{ash}{[0.134, 0.190]}}} &\makecell[t]{$0.111_{\textcolor{ash}{\pm 0.041}}$\\{\scriptsize \textcolor{ash}{[0.071, 0.125]}}} \\
 \hline
 
8 & Qwen & \makecell[t]{$0.712_{\textcolor{ash}{\pm 0.115}}$\\{\scriptsize \textcolor{ash}{[0.664, 0.761]}}} &\makecell[t]{$0.877_{\textcolor{ash}{\pm 0.033}}$\\{\scriptsize \textcolor{ash}{[0.829, 0.915]}}} &\makecell[t]{$0.709_{\textcolor{ash}{\pm 0.126}}$\\{\scriptsize \textcolor{ash}{[0.653, 0.767]}}} &\makecell[t]{$0.509_{\textcolor{ash}{\pm 0.262}}$\\{\scriptsize \textcolor{ash}{[0.418, 0.601]}}} &\makecell[t]{$0.161_{\textcolor{ash}{\pm 0.033}}$\\{\scriptsize \textcolor{ash}{[0.139, 0.186]}}} &\makecell[t]{$0.105_{\textcolor{ash}{\pm 0.067}}$\\{\scriptsize \textcolor{ash}{[0.072, 0.108]}}} \\
 & TabPFN & \makecell[t]{$0.751_{\textcolor{ash}{\pm 0.085}}$\\{\scriptsize \textcolor{ash}{[0.696, 0.802]}}} &\makecell[t]{$0.847_{\textcolor{ash}{\pm 0.044}}$\\{\scriptsize \textcolor{ash}{[0.779, 0.895]}}} &\makecell[t]{$0.756_{\textcolor{ash}{\pm 0.086}}$\\{\scriptsize \textcolor{ash}{[0.703, 0.813]}}} &\makecell[t]{$0.627_{\textcolor{ash}{\pm 0.142}}$\\{\scriptsize \textcolor{ash}{[0.539, 0.714]}}} &\makecell[t]{$0.170_{\textcolor{ash}{\pm 0.027}}$\\{\scriptsize \textcolor{ash}{[0.148, 0.196]}}} &\makecell[t]{$0.117_{\textcolor{ash}{\pm 0.031}}$\\{\scriptsize \textcolor{ash}{[0.069, 0.140]}}} \\
 & NODE & \makecell[t]{$0.753_{\textcolor{ash}{\pm 0.071}}$\\{\scriptsize \textcolor{ash}{[0.693, 0.803]}}} &\makecell[t]{$0.868_{\textcolor{ash}{\pm 0.043}}$\\{\scriptsize \textcolor{ash}{[0.811, 0.909]}}} &\makecell[t]{$0.753_{\textcolor{ash}{\pm 0.065}}$\\{\scriptsize \textcolor{ash}{[0.697, 0.807]}}} &\makecell[t]{$0.655_{\textcolor{ash}{\pm 0.184}}$\\{\scriptsize \textcolor{ash}{[0.570, 0.730]}}} &\makecell[t]{$0.153_{\textcolor{ash}{\pm 0.023}}$\\{\scriptsize \textcolor{ash}{[0.130, 0.179]}}} &\makecell[t]{$0.095_{\textcolor{ash}{\pm 0.033}}$\\{\scriptsize \textcolor{ash}{[0.074, 0.097]}}} \\
 & MLP & \makecell[t]{$0.752_{\textcolor{ash}{\pm 0.048}}$\\{\scriptsize \textcolor{ash}{[0.698, 0.801]}}} &\makecell[t]{$0.829_{\textcolor{ash}{\pm 0.038}}$\\{\scriptsize \textcolor{ash}{[0.766, 0.880]}}} &\makecell[t]{$0.751_{\textcolor{ash}{\pm 0.041}}$\\{\scriptsize \textcolor{ash}{[0.699, 0.804]}}} &\makecell[t]{$0.682_{\textcolor{ash}{\pm 0.144}}$\\{\scriptsize \textcolor{ash}{[0.599, 0.763]}}} &\makecell[t]{$0.171_{\textcolor{ash}{\pm 0.012}}$\\{\scriptsize \textcolor{ash}{[0.140, 0.208]}}} &\makecell[t]{$0.124_{\textcolor{ash}{\pm 0.028}}$\\{\scriptsize \textcolor{ash}{[0.081, 0.145]}}} \\
 \hline
 
16 & Qwen & \makecell[t]{$0.782_{\textcolor{ash}{\pm 0.051}}$\\{\scriptsize \textcolor{ash}{[0.729, 0.832]}}} &\makecell[t]{$0.878_{\textcolor{ash}{\pm 0.039}}$\\{\scriptsize \textcolor{ash}{[0.828, 0.917]}}} &\makecell[t]{$0.781_{\textcolor{ash}{\pm 0.052}}$\\{\scriptsize \textcolor{ash}{[0.732, 0.834]}}} &\makecell[t]{$0.736_{\textcolor{ash}{\pm 0.099}}$\\{\scriptsize \textcolor{ash}{[0.648, 0.816]}}} &\makecell[t]{$0.141_{\textcolor{ash}{\pm 0.029}}$\\{\scriptsize \textcolor{ash}{[0.116, 0.168]}}} &\makecell[t]{$0.091_{\textcolor{ash}{\pm 0.014}}$\\{\scriptsize \textcolor{ash}{[0.072, 0.091]}}} \\
 & TabPFN & \makecell[t]{$0.762_{\textcolor{ash}{\pm 0.042}}$\\{\scriptsize \textcolor{ash}{[0.713, 0.807]}}} &\makecell[t]{$0.859_{\textcolor{ash}{\pm 0.056}}$\\{\scriptsize \textcolor{ash}{[0.803, 0.901]}}} &\makecell[t]{$0.765_{\textcolor{ash}{\pm 0.037}}$\\{\scriptsize \textcolor{ash}{[0.718, 0.814]}}} &\makecell[t]{$0.655_{\textcolor{ash}{\pm 0.123}}$\\{\scriptsize \textcolor{ash}{[0.566, 0.734]}}} &\makecell[t]{$0.161_{\textcolor{ash}{\pm 0.030}}$\\{\scriptsize \textcolor{ash}{[0.136, 0.189]}}} &\makecell[t]{$0.141_{\textcolor{ash}{\pm 0.063}}$\\{\scriptsize \textcolor{ash}{[0.098, 0.162]}}} \\
 & NODE & \makecell[t]{$0.784_{\textcolor{ash}{\pm 0.092}}$\\{\scriptsize \textcolor{ash}{[0.735, 0.828]}}} &\makecell[t]{$0.872_{\textcolor{ash}{\pm 0.054}}$\\{\scriptsize \textcolor{ash}{[0.819, 0.912]}}} &\makecell[t]{$0.780_{\textcolor{ash}{\pm 0.091}}$\\{\scriptsize \textcolor{ash}{[0.732, 0.830]}}} &\makecell[t]{$0.727_{\textcolor{ash}{\pm 0.193}}$\\{\scriptsize \textcolor{ash}{[0.644, 0.802]}}} &\makecell[t]{$0.146_{\textcolor{ash}{\pm 0.031}}$\\{\scriptsize \textcolor{ash}{[0.124, 0.172]}}} &\makecell[t]{$0.106_{\textcolor{ash}{\pm 0.032}}$\\{\scriptsize \textcolor{ash}{[0.077, 0.111]}}} \\
 & MLP & \makecell[t]{$0.755_{\textcolor{ash}{\pm 0.056}}$\\{\scriptsize \textcolor{ash}{[0.701, 0.805]}}} &\makecell[t]{$0.830_{\textcolor{ash}{\pm 0.071}}$\\{\scriptsize \textcolor{ash}{[0.771, 0.875]}}} &\makecell[t]{$0.751_{\textcolor{ash}{\pm 0.070}}$\\{\scriptsize \textcolor{ash}{[0.700, 0.805]}}} &\makecell[t]{$0.727_{\textcolor{ash}{\pm 0.064}}$\\{\scriptsize \textcolor{ash}{[0.643, 0.803]}}} &\makecell[t]{$0.175_{\textcolor{ash}{\pm 0.044}}$\\{\scriptsize \textcolor{ash}{[0.145, 0.209]}}} &\makecell[t]{$0.153_{\textcolor{ash}{\pm 0.034}}$\\{\scriptsize \textcolor{ash}{[0.111, 0.179]}}} \\
 \hline
 
32 & Qwen & \makecell[t]{$0.764_{\textcolor{ash}{\pm 0.055}}$\\{\scriptsize \textcolor{ash}{[0.713, 0.810]}}} &\makecell[t]{$0.874_{\textcolor{ash}{\pm 0.031}}$\\{\scriptsize \textcolor{ash}{[0.828, 0.912]}}} &\makecell[t]{$0.770_{\textcolor{ash}{\pm 0.051}}$\\{\scriptsize \textcolor{ash}{[0.720, 0.819]}}} &\makecell[t]{$0.636_{\textcolor{ash}{\pm 0.129}}$\\{\scriptsize \textcolor{ash}{[0.543, 0.721]}}} &\makecell[t]{$0.146_{\textcolor{ash}{\pm 0.023}}$\\{\scriptsize \textcolor{ash}{[0.122, 0.173]}}} &\makecell[t]{$0.095_{\textcolor{ash}{\pm 0.029}}$\\{\scriptsize \textcolor{ash}{[0.068, 0.096]}}} \\
 & TabPFN & \makecell[t]{$0.820_{\textcolor{ash}{\pm 0.041}}$\\{\scriptsize \textcolor{ash}{[0.769, 0.865]}}} &\makecell[t]{$0.871_{\textcolor{ash}{\pm 0.046}}$\\{\scriptsize \textcolor{ash}{[0.818, 0.910]}}} &\makecell[t]{$0.825_{\textcolor{ash}{\pm 0.036}}$\\{\scriptsize \textcolor{ash}{[0.779, 0.871]}}} &\makecell[t]{$0.736_{\textcolor{ash}{\pm 0.099}}$\\{\scriptsize \textcolor{ash}{[0.643, 0.815]}}} &\makecell[t]{$0.134_{\textcolor{ash}{\pm 0.028}}$\\{\scriptsize \textcolor{ash}{[0.113, 0.158]}}} &\makecell[t]{$0.105_{\textcolor{ash}{\pm 0.049}}$\\{\scriptsize \textcolor{ash}{[0.075, 0.113]}}} \\
 & NODE & \makecell[t]{$0.797_{\textcolor{ash}{\pm 0.075}}$\\{\scriptsize \textcolor{ash}{[0.753, 0.841]}}} &\makecell[t]{$0.871_{\textcolor{ash}{\pm 0.065}}$\\{\scriptsize \textcolor{ash}{[0.818, 0.910]}}} &\makecell[t]{$0.801_{\textcolor{ash}{\pm 0.071}}$\\{\scriptsize \textcolor{ash}{[0.758, 0.847]}}} &\makecell[t]{$0.709_{\textcolor{ash}{\pm 0.153}}$\\{\scriptsize \textcolor{ash}{[0.627, 0.785]}}} &\makecell[t]{$0.142_{\textcolor{ash}{\pm 0.033}}$\\{\scriptsize \textcolor{ash}{[0.122, 0.166]}}} &\makecell[t]{$0.097_{\textcolor{ash}{\pm 0.035}}$\\{\scriptsize \textcolor{ash}{[0.075, 0.097]}}} \\
 & MLP & \makecell[t]{$0.771_{\textcolor{ash}{\pm 0.145}}$\\{\scriptsize \textcolor{ash}{[0.719, 0.817]}}} &\makecell[t]{$0.814_{\textcolor{ash}{\pm 0.131}}$\\{\scriptsize \textcolor{ash}{[0.757, 0.862]}}} &\makecell[t]{$0.768_{\textcolor{ash}{\pm 0.143}}$\\{\scriptsize \textcolor{ash}{[0.719, 0.817]}}} &\makecell[t]{$0.736_{\textcolor{ash}{\pm 0.202}}$\\{\scriptsize \textcolor{ash}{[0.664, 0.803]}}} &\makecell[t]{$0.160_{\textcolor{ash}{\pm 0.055}}$\\{\scriptsize \textcolor{ash}{[0.133, 0.192]}}} &\makecell[t]{$0.140_{\textcolor{ash}{\pm 0.066}}$\\{\scriptsize \textcolor{ash}{[0.106, 0.155]}}} \\
 \hline

 \hline
    \end{tabular}
}
\end{table}

For conventional ML models, the pairwise comparisons show a wider spread. MLP and stronger ensemble models form the leading group, whereas simpler or less stable models, such as LR and DT, tend to show higher loss. The detailed metrics in Table~\ref{tab:top4_metrics} show that MLP remains a competitive supervised baseline, but its Brier loss is generally higher than the strongest LLM and tabular foundation-model baselines, ranging from 0.160 to 0.175 across training sizes. This indicates that although MLP achieves competitive classification performance, its probabilistic calibration is weaker than Qwen-3 in low-shot settings and TabPFN at larger training sizes.

The cross-family comparison in Fig.~\ref{fig:meanloss}d shows that the best LLM, Qwen-3, is competitive with the strongest ML and tabular DL/foundation-model baselines. Table~\ref{tab:top4_metrics} clarifies this trend: Qwen-3 is particularly strong when only a few in-context examples are available, whereas TabPFN and NODE become more competitive as labeled training data increase. Overall, selected-feature LLM prompting provides strong low-resource probabilistic performance, while supervised tabular models retain an advantage when sufficient labeled samples are available for training.

\begin{table}[!t]
    \centering
    \caption{Comparison of feature importance rankings and epidemiological associations. The table reports an ML-derived reference ranking from the MLP model and rankings induced by LLM-based surrogate models. For each feature, the related risk ratio (RR) with its 95\% CI is shown, along with the estimation type (standard or penalized). 
    }
    \label{tab:f-ranking}
    \setlength{\tabcolsep}{3pt} 
    \resizebox{\linewidth}{!}{
    \begin{tabular}{l@{\extracolsep{-4pt}}cccccc@{\extracolsep{2pt}} ccl}
    \hline
       Feature  & \multicolumn{6}{c}{Feature ranking} & \multicolumn{3}{c}{Epidemiological associations}\\ \cline{2-7}\cline{8-10}
       & ML Reference & \multicolumn{5}{c}{LLM-induced surrogate} & Related & 95\% CI & RR type\\ \cline{3-7}
       & MLP & Gemma & GPT & Llama & Mistral & Qwen & risk (RR) & &\\
       \hline\hline
        Hypertension &	1 &	3 & 3 & 3 & 1 & 3 & 4.03* & [2.97, 5.48]	& Standard\\
        Age\_\textit{60+y} & 2 & 1 & 5 & 5 & 3 & 4 & 3.68*$\dagger$ & [2.98, 4.55] & Penalized\\
        RBC & 3	& 4 & 6 & 6 & 9 & 6 & 2.80*$\dagger$ & [2.34, 3.34] & Penalized\\
        Daily sleep $<7$h & 4 & 8 & 7 & 8 & 4 & 8 & 1.75* & [1.33, 2.30] & Standard\\
        Anemia & 8 & 5 & 2 & 2 & 5 & 5 & 1.64* & [1.24, 2.17] & Standard\\
        Diabetes & 7 & 2 & 1 & 4 & 2 & 1 & 1.56* & [1.16, 2.09] & Standard\\
        BMI\_\textit{Obese} &5 & 7 & 4 & 1 & 6 & 2 & 1.47* & [1.10, 1.94] & Standard\\
        Family hypertension & 9 & 9 & 9 & 9 & 8 & 7 & 1.08 & [0.80, 1.45] & Standard\\
        Gender\_\textit{Male} & 6 & 6 & 8 & 7 & 7 & 9 & 0.77 & [0.57, 1.05] & Standard\\
\hline
\multicolumn{9}{l}{* 95\% confidence interval excluding 1.0 ($p < 0.05$)}\\
\multicolumn{9}{l}{$\dagger$ penalized estimate computed due to perfect separation (absence of non-CKD cases)}
    \end{tabular}
}
\end{table}

\subsection{Feature Ranking and Clinical Plausibility}
\label{sec:feature_ranking_results}
Table~\ref{tab:f-ranking} compares the MLP-based ML reference ranking with LLM-derived feature rankings and epidemiological risk-ratio associations. The ML reference ranking emphasizes clinically established CKD risk factors, placing hypertension, age $\geq 60$, and RBC as the top three features. These variables also show strong epidemiological associations, with elevated risk ratios for CKD. This agreement suggests that the MLP reference captures major data-driven and clinically plausible risk patterns in the study cohort.

LLMs show partially overlapping but distinct ranking behavior. Several models assign high importance to diabetes, anemia, and obesity-related features, which are clinically meaningful CKD risk factors but ranked lower by the ML reference. Diabetes is ranked first by Llama-3 and Qwen-3 and second by GPT-4o-mini, while BMI\_\textit{Obese} is ranked highest by Gemma-2. Hypertension remains highly ranked by most models, especially Mistral, which assigns it the top rank. In contrast, RBC is emphasized strongly by the ML reference but receives moderate or lower rankings from most LLMs, suggesting that some LLMs may underweight urinalysis-related evidence relative to broader comorbidity patterns.

\begin{figure}[!t]
    \centering
    \includegraphics[width=.8\linewidth]{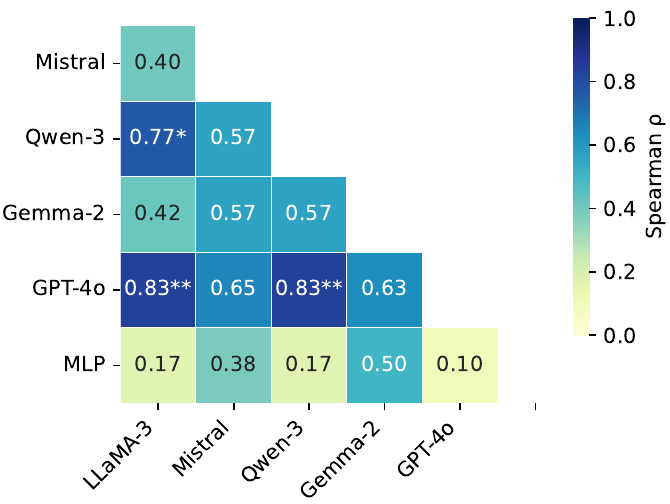}
    \caption{Spearman rank correlation heatmap comparing feature-importance rankings among large language models and the MLP model. Each cell reports the Spearman correlation coefficient ($\rho$) between a pair of models; darker colors indicate stronger agreement. 
    }
    \label{fig:feature-spearman}
\end{figure}

Fig.~\ref{fig:feature-spearman} further quantifies ranking agreement using Spearman correlation. LLM rankings show moderate to strong agreement with each other, particularly among GPT-4o-mini, Qwen-3, and Llama-3, indicating that LLMs tend to follow a shared feature-prioritization pattern. However, correlations between the ML reference ranking and LLM rankings are generally low, showing that LLMs do not simply reproduce the supervised tabular importance structure. Among the LLMs, Gemma-2 shows the highest agreement with the ML reference, while GPT-4o-mini shows weaker agreement.

Overall, these results indicate that LLMs prioritize clinically meaningful CKD risk factors but often differ from the MLP-based ML reference in how they weight those factors. The ML reference places greater emphasis on cohort-specific predictors such as hypertension, age, and urinary RBC, whereas LLMs more strongly emphasize broadly recognized clinical risk factors such as diabetes, anemia, and obesity. This suggests that LLM-based CKD screening is clinically plausible, but its feature-prioritization pattern is not identical to supervised tabular learning and should be interpreted with caution.

\begin{table}[!t]
    \centering
        \caption{Comparison of zero-shot LLM performance with established CKD screening tools on the full Dataset-1 cohort. Balanced accuracy (Bal. Acc.) is reported for both the LLMs and tools. Reported values are mean Brier loss differences (LLM$-$Tool), where negative values indicate better probabilistic performance of the LLM. Statistical significance is assessed using a paired permutation test on per-sample Brier loss. 
        }
    \label{tab:llm-vs-tools}
    \setlength{\tabcolsep}{1pt} 
    \resizebox{\linewidth}{!}{
    \begin{tabular}{lc|ccccc|}
    \cline{3-7}
      & Tool$\rightarrow$  & SCORED~\cite{bang2007development} & Kshirsagar~\cite{kshirsagar2008simple} & Thakkinstian~\cite{thakkinstian2011simple} & Kwon~\cite{kwon2019simple} & Kearns~\cite{kearns2013predicting} \\
    LLM   &Bal. Acc.  & 0.7635 & 0.7709 & 0.8103 & 0.7749 & 0.7829\\
   \hline
     \multicolumn{1}{|l}{Qwen} & 0.7829 & \hm{-0.150}{***} & \hm{-0.111}{***} & \hm{-0.120}{***} & \hm{-0.206}{***} & \hm{-0.116}{***} \\
    \multicolumn{1}{|l}{Mistral}& 0.7925 & \hm{-0.148}{***} & \hm{-0.110}{***} & \hm{-0.119}{***} & \hm{-0.205}{***} & \hm{-0.114}{***} \\
    \multicolumn{1}{|l}{Llama} & 0.7422 & \hm{-0.065}{**} & \hm{-0.027}{n.s} & \hm{-0.036}{n.s} & \hm{-0.122}{***} & \hm{-0.031}{n.s} \\    
    \multicolumn{1}{|l}{GPT-4o} & 0.7725 & \hm{-0.098}{***} & \hm{-0.059}{*} & \hm{-0.068}{**} & \hm{-0.154}{***} & \hm{-0.064}{**} \\           
    \multicolumn{1}{|l}{Gemma} & 0.7299 & \hm{-0.134}{***} & \hm{-0.095}{***} & \hm{-0.104}{***} & \hm{-0.190}{***} & \hm{-0.099}{***} \\       
    
    \hline
       
    \end{tabular}
    }
\end{table}

\subsection{Comparison with CKD Screening Tools}
\label{sec:screening_tools}

Table~\ref{tab:llm-vs-tools} compares zero-shot LLM performance with established CKD screening tools on the full Dataset-1 cohort. The screening tools achieve balanced accuracy between 0.7635 and 0.8103, with Thakkinstian obtaining the highest balanced accuracy among the rule-based tools. Pairwise values report mean Brier loss differences, defined as $\Delta=\text{Loss}(\text{LLM})-\text{Loss}(\text{Tool})$; thus, negative values indicate better probabilistic performance for the LLM.

Qwen-3 and Mistral provide the greatest and most consistent probabilistic improvements over the screening tools. Qwen-3 achieves balanced accuracy of 0.7829, matching or exceeding most tools except Thakkinstian~\cite{thakkinstian2011simple}, while producing significantly lower Brier loss than all five tools, with differences ranging from $-0.111$ to $-0.206$ ($p<0.001$). Mistral obtains the highest LLM balanced accuracy of 0.7925 and shows a nearly identical pattern, with significant Brier loss reductions across all tool comparisons. These results suggest that the best zero-shot LLMs can provide competitive classification performance while achieving lower Brier loss than established rule-based tools.

GPT-4o-mini and Gemma-2 also improve probabilistic performance over most or all tools, although their balanced accuracy is lower than that of Qwen-3 and Mistral. GPT-4o-mini achieves a balanced accuracy of 0.7725 and significantly reduces Brier loss against all tools, with the largest reduction observed against Kwon~\cite{kwon2019simple}. Gemma-2 has a lower balanced accuracy of 0.7299, but still yields significant Brier loss reductions across all comparisons, indicating that classification accuracy and probabilistic quality do not always move together. In contrast, Llama-3 shows weaker and less consistent gains: it significantly outperforms SCORED~\cite{bang2007development} and Kwon~\cite{kwon2019simple} in Brier loss, but its differences against Kshirsagar~\cite{kshirsagar2008simple}, Thakkinstian~\cite{thakkinstian2011simple}, and Kearns~\cite{kearns2013predicting} are small and non-significant.

Taken together, these results show that zero-shot LLMs can be competitive with established CKD screening tools in both classification and probabilistic performance. Qwen-3 and Mistral offer the strongest overall trade-off between balanced accuracy and Brier loss, whereas Llama-3 highlights that this advantage is model-dependent rather than universal across LLMs.

\begin{figure}[!t]
\centering
\begin{tikzpicture}
\begin{groupplot}[
    enlargelimits=0.15,
    footnotesize,
    group style={
        group size=2 by 1,
        xlabels at=edge bottom,
        horizontal sep=.7cm,   
        ylabels at=edge left
    }
]
\nextgroupplot[
    ybar,
    nodes near coords,
    every node near coord/.append style={rotate=90,anchor=west, /pgf/number format/precision=2,  font=\scriptsize}, 
    width=0.3\textwidth,
    height=5cm,
    bar width=6pt,
    ylabel={Balanced Accuracy},
    xlabel={\shortstack{\\\\\\\\(a)}},
    ymin=0.44, ymax=1.1,
    axis y discontinuity=crunch,
    ytick={0.5,0.6,0.7,0.8,0.9},
    yticklabels={0.5,0.6,0.7,0.8,0.9},
    xtick={0,1,2,3,4},
    xticklabel style={rotate=45, anchor=east, align=center},
     xticklabels={Gemma-2,Llama-3,Qwen-3,Mistral, GPT-4o},
    legend style={at={(0.45,-0.35)},font = \scriptsize, anchor=north,legend columns=2},
]

\addplot[pattern=dots, pattern color=SciRed, draw=SciRed, error bars/.cd, y dir=both, y explicit] 
    coordinates { (0, 0.896)
    (1,0.73429) (2, 0.81605) (3, 0.86487) (4,0.85567)};
\addlegendentry{All features}
\addplot[pattern=north east lines, pattern color=SciBlue, draw=SciBlue,  error bars/.cd, 
          y dir=both, y explicit] 
    coordinates {      
        (0, 0.824) 
        (1, 0.88309) 
        (2, 0.81237) 
        (3, 0.88609) 
        (4, 0.88001) 
        };
\addlegendentry{Selected features}
\sigbar{0}{1.06}{0.04}{***} 
\sigbar{1}{1.06}{0.04}{**} 
\sigbar{2}{1.06}{0.04}{**} 
\sigbar{3}{1.06}{0.04}{*}
\sigbar{4}{1.06}{0.04}{$n.s$}

\nextgroupplot[
    xlabel={\shortstack{Shots\\\\\\\\\\\\\\\\\\\\\\\\(b)}},
    width=0.25\textwidth,
    height=5cm,
    ymin=0.81, ymax=0.94,
    symbolic x coords={0,4,8,16,32},
    xtick=data,
    legend style={at={(0.5,-0.25)},font=\scriptsize,
    /tikz/every even column/.append style={column sep=1pt},
    row sep=-2pt, anchor=north,legend columns=2}
]
\addplot[color=SciRed, mark=*,thick]
coordinates {(0,0.8440) (4,0.8920) (8,0.8980) (16,0.9067) (32,0.918)};
\addlegendentry{Gemma-2}

\addplot[color=SciOrange,mark=+,thick]
coordinates {(0,0.892) (4,0.882) (8,0.888) (16,0.8913) (32,0.9207)};
\addlegendentry{GPT-4o}

\addplot[color=cyan,mark=x, thick]
coordinates {(0,0.898) (4,0.858) (8,0.8893) (16,0.904) (32,0.8413)};
\addlegendentry{Llama-3}

\addplot[color=SciGreen, mark=diamond*, thick]
coordinates {(0,0.824) (4,0.9) (8,0.892) (16,0.918) (32,0.9233)};
\addlegendentry{Qwen-3}

\addplot[ color=gray, mark=triangle*,thick]
coordinates {(0,0.9) (4,0.892) (8,0.888) (16,0.9) (32,0.912)};
\addlegendentry{Mistral}

\addplot[color=CustomPurple, mark=pentagon*,thick]
coordinates {(4,0.852) (8,0.8987) (16,0.914) (32,0.946) };
\addlegendentry{TabPFN}


\end{groupplot}
\end{tikzpicture}
\caption{Independent-dataset evaluation on Dataset-2. (a) Balanced accuracy for \textit{all} versus selected features; significance is assessed using a paired permutation test on per-sample Brier loss. (b) Few-shot performance across shot settings, comparing LLMs with TabPFN as a supervised tabular baseline. 
$n.s$ means `not significant' ($p \geq 0.05$).
}
\label{fig:externalValidation}
\end{figure}
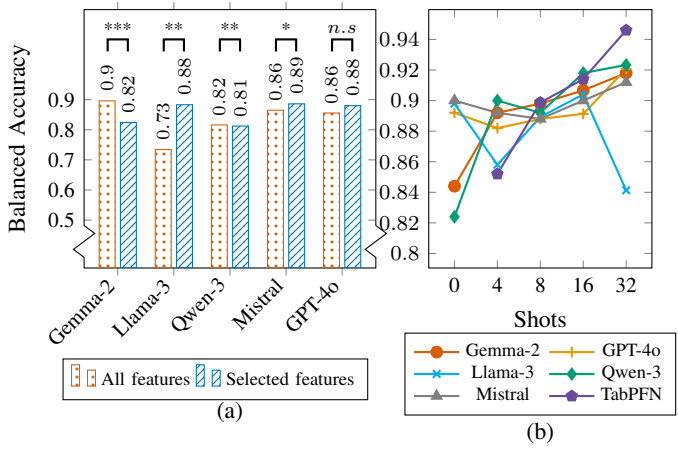

\subsection{Independent Dataset Evaluation}
\label{sec:cross_dataset}

Fig.~\ref{fig:externalValidation} evaluates model performance on the independent Dataset-2 cohort. Fig.~\ref{fig:externalValidation}a compares all-feature and selected-feature zero-shot prompting. Feature selection improves or maintains performance for most models, but the effect is model-dependent. Llama-3 shows the largest gain, increasing from 0.73 to 0.88 balanced accuracy ($p<0.01$). Mistral and GPT-4o-mini also improve slightly, from 0.86 to 0.89 and 0.86 to 0.88, respectively, although these differences are not statistically significant. Qwen-3 remains stable, with a small decrease from 0.82 to 0.81 ($p<0.01$). In contrast, Gemma-2 decreases from 0.90 to 0.82 ($p<0.001$), indicating sensitivity to feature reduction under distribution shift.

Fig.~\ref{fig:externalValidation}b examines performance across increasing numbers of in-context examples using selected features, with TabPFN included as a supervised tabular baseline. LLM performance is generally strong on Dataset-2, with balanced accuracy mostly between 0.84 and 0.92. Gemma-2 and Qwen-3 improve with additional examples, reaching approximately 0.92 at higher shots. Mistral and GPT-4o-mini also improve gradually and approach 0.91--0.92 at 32 shots. Llama-3 is less stable, declining at higher shots. TabPFN shows the most consistent scaling behavior, increasing from approximately 0.85 with 4 context samples to 0.94 with 32 samples.
Because Dataset-2 is hospital-based and does not report CKD stage, these results provide preliminary evidence of performance across datasets and do not validate early-stage community screening.

\subsection{Limitations}
\label{sec:limitations}
This study has several limitations. First, the primary cohort is relatively small, which may limit the stability of performance estimates. Although repeated runs and an independent-dataset evaluation were used to reduce this concern, larger and more diverse cohorts are needed for stronger generalization claims. Second, LLM reliability depends on prompt design, feature serialization, model version, and availability of token-level log-probabilities; therefore, results may vary across future model releases and deployment settings. Third, the selected feature subset was derived from prior ML-based analysis and may not be optimal for all populations or clinical contexts. Fourth, the independent dataset differs from the primary cohort in sampling setting, feature availability, and case-control composition; therefore, these results provide only preliminary evidence of cross-dataset performance rather than definitive clinical validation. 
Finally, we did not directly evaluate robustness to patient-level missing features, heterogeneous inputs, deployment efficiency, clinical safety, or cost; these factors, together with prospective validation, require evaluation before clinical use.

\section{Conclusion}

This paper presented \textit{LLM4CKD}, a comprehensive applied evaluation of large language models for low-resource early-stage CKD screening under zero-shot and few-shot in-context learning. We benchmarked five LLMs against conventional ML, tabular DL, tabular foundation-model baselines, established CKD screening tools, and an independent dataset. The results show that LLMs can provide competitive CKD screening with minimal labeled data, particularly when clinically compact selected-feature prompts are used. Among the evaluated models, Qwen-3-8B, Mistral, and GPT-4o-mini achieved strong balanced accuracy and probabilistic performance, and the strongest zero-shot LLMs improved Brier loss relative to established rule-based screening tools. At the same time, additional in-context examples did not consistently improve performance, indicating that LLM-based screening remains sensitive to model choice, prompt design, and example composition. Supervised ML and tabular models improved more steadily as labeled training data increased, highlighting their continued advantage when sufficient training data and training infrastructure are available.

The feature-ranking analysis showed that LLMs emphasize clinically meaningful CKD risk factors, but their prioritization patterns differ from the MLP-based ML reference, indicating distinct feature prioritization between the two approaches. Evaluation on an independent dataset provided preliminary evidence of performance across a different cohort and data setting, although broader external validation is required. Overall, \textit{LLM4CKD} shows that LLMs may support low-data CKD screening and probabilistic prediction, while supervised tabular models remain important when local labeled data are available. Future work should evaluate robustness to missing and heterogeneous inputs, computational efficiency, and prospective performance across broader clinical settings.

{
\bibliographystyle{ieee_fullname}
\bibliography{reference}
}

\end{document}